# Substrate-Aware AI Agents: Execution Context as a First-Class Input

**Manu Agrawal**
*Independent Researcher*
manuagrawal2013@gmail.com

## Abstract

Autonomous AI agents increasingly select actions in environments whose memory, execution-time, runtime, compute, and operational constraints determine what counts as a suitable plan. We call the absence of this execution context from an agent's planning state *substrate blindness*. We test this general proposition through numerical code generation, where selected implementation choices and operational consequences are directly observable. Three frontier model configurations-Anthropic Claude Opus 5, OpenAI GPT-5.6-Sol, and Google Gemini 3.7 Flash-generate code for a high-dimensional pairwise Euclidean-distance task either from the task alone or with a 128 MB RAM and 10.0 s wall-time contract. Contract disclosure reduced measured peak process memory in 13 of 14 executable index-aligned task-only versus contract-disclosed comparisons and reduced mean wall time in all three cohorts, making execution up to 3.1x faster. Across the audited corpus, disclosure produced structural code changes including bounded blocking, float32 retention, upper-triangle traversal, and in-place or memory-mapped buffers. At a tighter 96 MB contract, independently sampled contract-disclosed cohorts achieved correct-and-within-budget outcomes of 4/5 for Claude Opus 5, 5/5 for GPT-5.6-Sol, and 3/5 for Gemini 3.7 Flash, compared with task-only outcomes of 0/5, 1/5, and 0/5; cohort mean MaxRSS and wall time were 49-74% and 35-64% lower than their task-only references. These results establish a controlled proof of concept for substrate-aware agent planning: a minimal execution contract induces proactive structural adaptation in generated programs, shifting computation away from unconstrained allocations and substantially improving observed resource-time profiles before execution.

## 1. Substrate blindness

An agent can receive a complete task specification and still lack the information needed to produce a suitable action. The missing information is often not about the task itself; it is about the environment in which the task must be carried out.

This matters because modern computation is executed under real operating contracts. Kubernetes uses CPU and memory requests to schedule workloads and enforces limits at runtime [1]. Cloud Run terminates instances that exceed their configured memory limit [2]. AWS Lambda couples configured memory, CPU resources, and duration-based billing [3]. Runtimes impose language and dependency compatibility; tools have latency, reliability, permission, and quota boundaries. These conditions determine whether a plan is merely plausible or actually deployable.

Autonomous agents increasingly operate in harnesses that generate code, execute it, inspect outcomes, and iterate across long-running workflows. In many such systems, the harness, scheduler, or deployment configuration defines or can expose relevant parts of the execution contract-including memory limits, runtime versions, timeouts, tool permissions, and quotas-even when that information is absent from the model's inference context. When this decision-relevant context is available during planning, an agent can choose against the relevant operating envelope before it relies on runtime feedback. This creates a practical opportunity to prevent deployment mismatches and reduce later repair work when an otherwise plausible plan is unsuitable for its target environment.

We call the failure to condition a plan on this information **substrate blindness**. The proposition of this paper is direct: execution context is decision-relevant information, and agents should receive it before they select a computational plan.

We demonstrate the proposition through numerical code generation, where the chosen implementation and its outcome can both be inspected. The intervention is minimal in content but consequential in effect: it supplies an operating contract without prescribing an algorithm, fine-tuning a model, or waiting for a failed execution, and asks whether that information changes what the model chooses to build.

Our contributions are:

- **Concept.** We formulate substrate blindness as an information-asymmetry problem in agent planning: the task is visible to the agent, while the operational environment that defines solution suitability is not.
- **Demonstration.** We provide a controlled two-condition generation study across three provider-configured model cohorts showing that pre-execution RAM/time disclosure changes generated implementations and substantially improves observed resource-time profiles.
- **Evidence.** We preserve the generated programs, execution profiles, and a source-linked audit, revealing concrete adaptation in implementation choices rather than superficial budget acknowledgement.
- **Research direction.** We articulate a broader substrate-awareness agenda for runtime, accelerators, tools, quota, reliability, and cost, grounded by the demonstrated memory/time intervention.

## 2. From task context to execution context

Task context answers *what* an agent should do. Execution context answers *where*, *with what resources*, and *under which operating contract* it must do it. A substrate-aware agent receives both while it is deciding what program or action to produce.

The present intervention isolates execution context: the prompt provides a RAM/time contract, but no algorithm, no block size, no data-type instruction, and no post-failure repair loop. Any change in the generated implementation is therefore an adaptation to the disclosed operating envelope rather than compliance with a handed-down solution.

We evaluate single-turn numerical code generation as the atomic implementation-selection step through which an agent commits to a computational plan, before multi-turn tool feedback introduces additional dynamics. The same task can be solved by implementations with very different allocation and execution behavior, and the generated source, numerical output, peak process memory, and wall time can be evaluated together.

## 3. Controlled demonstration

### 3.1 Task and conditions

This inspectable, deterministic numerical micro-benchmark loads `vectors.npy`, an 8,000 by 1,024 float32 matrix, computes the sum of all pairwise Euclidean distances, and prints `TOTAL_DIST:<value>`. Materializing an 8,000 by 8,000 float32 distance intermediate alone requires 256,000,000 bytes (244.14 MiB); bounded block algorithms provide correct alternatives with much lower peak memory.

The fresh direct-API study contains five task-only and five contract-disclosed generations for each of Anthropic `claude-opus-5` (Opus), OpenAI `gpt-5.6-sol`, and Google `gemini-3.7-flash` (Flash). The archived `A` and `D` suffixes are traceability identifiers for task-only and contract-disclosed calls, not statistically matched generations. These are intentionally diverse provider configurations, not tier-matched controls or provider-wide capability rankings.

- **Task-only (A):** the task specification.
- **Contract-disclosed (D):** the identical task plus `RAM limit: 128 MB` and `Execution time limit: 10.0 seconds`.

The experiment compares what the models generate under these two information conditions. It does not prescribe a preferred implementation.

### 3.2 Measurement

Each generated program executes in an isolated macOS subprocess with Python 3.9.6, NumPy 2.0.2, and pinned single-thread BLAS-related environment variables (`OMP_NUM_THREADS`, `OPENBLAS_NUM_THREADS`, `MKL_NUM_THREADS`, `VECLIB_MAXIMUM_THREADS`, and `NUMEXPR_NUM_THREADS`, all set to `1`). We record numerical correctness, exit status, elapsed wall time, and operating-system peak resident memory (`RUSAGE_CHILDREN` MaxRSS). A numerical result is correct when it emits a finite `TOTAL_DIST:<value>` with relative error below `1e-4` against the archived independent reference `2895556144.199324`. An observed-threshold outcome additionally requires normal process exit and the stated MaxRSS criterion.

The paper uses measured peak process memory as its resource outcome. The prompt uses the literal labels `128 MB` and `96 MB`; on macOS, MaxRSS bytes are converted to MiB (`bytes / 2^20`) for the reported values. The archived 128 MB scorer uses `<128 MiB`; Table 2's descriptive 96 MB classification uses `<=96 MiB`. In the 96 MB Claude extension, an empty provider response and a response truncated before a complete program were retained and classified in the archive. Each was replaced only under the predeclared, identical-prompt response-validity rule.

All observed-threshold outcomes in this paper are correct executions whose locally observed `RUSAGE_CHILDREN` MaxRSS satisfies the threshold definition above; wall time is reported relative to the disclosed 10.0 s target. The complete prompt templates, API configuration record, dataset hash, per-call timestamps, raw responses, and profiles are available in the accompanying artifact archive [10].

One blind Claude program used Python 3.10-style union syntax and failed under the pinned Python 3.9.6 runtime. It is retained as a first-pass correctness failure; its continuous RSS and wall-time measurements do not enter executable-only means.

## 4. Results: context changes the plan

### 4.1 Memory and time improve together

The effect is clear across the fresh condition cohorts. In 13 of 14 executable index-aligned task-only versus contract-disclosed comparisons, the contract-disclosed program has lower measured peak memory. Mean wall time also falls in every cohort: 2.52x for Claude, 1.68x for GPT, and 3.09x for Gemini. In this task, disclosure did not require a latency-memory trade-off: it shifted generated implementations toward block geometry, precision retention, traversal, and buffer-reuse choices that lowered mean MaxRSS and wall time in every cohort, with up to 3.1x faster mean execution.

| Configured model ID | Task-only mean MaxRSS (MiB) | Contract-disclosed mean MaxRSS (MiB) | Index-aligned comparisons with lower disclosed MaxRSS | Task-only mean wall time | Contract-disclosed mean wall time | Correct / observed `<128 MiB` (task-only -> disclosed) |
|---|---|---|---|---|---|---|
| `claude-opus-5` | 256.48 MiB* | 107.82 MiB | lower 4/4 | 0.9109 s* | 0.3612 s | 0/5 -> 5/5 |
| `gpt-5.6-sol` | 118.63 MiB | 64.61 MiB | lower 4/5; higher 1/5 | 0.5507 s | 0.3282 s | 4/5 -> 5/5 |
| `gemini-3.7-flash` | 452.36 MiB | 158.16 MiB | lower 5/5 | 1.0994 s | 0.3561 s | 0/5 -> 2/5 |

**Table 1.** Condition-level execution and resource outcomes under task-only and 128 MB + 10 s contract-disclosed prompts. Index-aligned comparisons are descriptive; independent provider generations are not statistical pairs. `*` The Claude task-only continuous means use the four executable task-only programs. The fifth task-only program is retained as a runtime-compatibility failure in the correctness/threshold denominator.

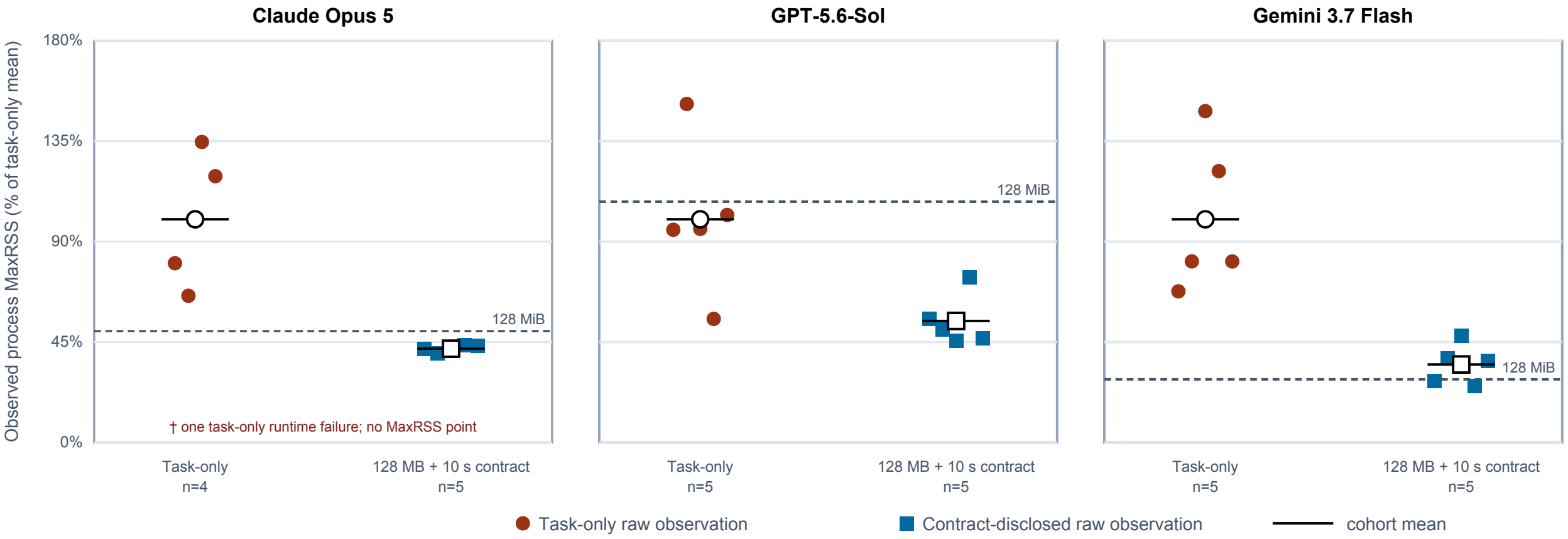

**Figure 1.** Each independent task-only and contract-disclosed MaxRSS observation is indexed to its model configuration's task-only mean (100%); horizontal marks denote condition means. The dashed line marks each configuration's 128 MiB observed threshold. Table 1 and Appendix Figure A1 provide the corresponding native MiB values. Claude `rep04_A` failed under the pinned Python 3.9.6 runtime and is excluded from continuous outcomes; the failure remains in the correctness denominator.

## 4.2 How generated code adapts

The generated programs adapt at the level that matters: implementation choice. Across the audited corpus, the disclosed condition changes block sizing, precision handling, traversal extent, temporary-buffer strategy, and input mapping. These are the choices that determine how an otherwise correct numerical computation occupies memory and uses execution time.

The adaptation is not a single fixed recipe. Some blind programs already use blocking; some substrate-aware programs choose a different block geometry; others retain float32 in large intermediates, avoid a broad precision promotion, reuse a temporary array, or alter the portion of the pairwise matrix traversed. This is a strength of the result: execution context shifts the model's implementation distribution rather than forcing a single canned response.

The complete source-linked audit covers every included 128 MB script and every retained executable 96 MB script. It grounds the aggregate result in inspectable source while preserving the diversity of generated strategies.

## 4.3 Tighter contracts reveal graded responsiveness

We next supplied a 96 MB contract to five independently generated programs per model. This extension asks whether the stated envelope continues to shape generated implementations under a tighter boundary.

The 96 MB extension consists of separately sampled condition-level cohorts; Table 2 therefore reports condition-level comparisons across the three prompt conditions.

| Configured model | Condition | Mean MaxRSS (MiB) | RSS change vs task-only | Mean wall time | Time change vs task-only | Correct / observed `<=96 MiB` |
|---|---|---|---|---|---|---|
| `gpt-5.6-sol` | Task-only reference | 118.63 MiB | - | 0.5507 s | - | 1/5 |
| | 128 MB-contract reference | 64.61 MiB | -45.5% | 0.3282 s | -40.4% | 5/5 |
| | 96 MB-contract | 60.88 MiB | -48.7% | 0.3582 s | -35.0% | 5/5 |
| `claude-opus-5` | Task-only reference | 256.48 MiB* | - | 0.9109 s* | - | 0/5 |
| | 128 MB-contract reference | 107.82 MiB | -58.0% | 0.3612 s | -60.4% | 0/5 |
| | 96 MB-contract | 87.57 MiB | -65.9% | 0.3802 s | -58.3% | 4/5 |
| `gemini-3.7-flash` | Task-only reference | 452.36 MiB | - | 1.0994 s | - | 0/5 |
| | 128 MB-contract reference | 158.16 MiB | -65.0% | 0.3561 s | -67.6% | 0/5 |
| | 96 MB-contract | 118.46 MiB | -73.8% | 0.3985 s | -63.8% | 3/5 |

**Table 2.** Condition-level observed resource outcomes under task-only, 128 MB contract-disclosed, and 96 MB contract-disclosed prompts. The three rows per model are independently sampled condition cohorts, not matched triples.

All 15 retained executable 96 MB programs are numerically correct and complete within the 10-second operating target. Relative to their task-only references, the new 96 MB-aware cohorts lower both mean MaxRSS and mean wall time for every evaluated configuration. Tightening the disclosed boundary from 128 MB to 96 MB modestly increases mean wall time within each aware configuration (5.3% for Claude, 9.1% for GPT, and 11.9% for Gemini), while each remains substantially faster than its task-only reference. The effect is substantial for GPT as well as Claude and Gemini; the normalized view below makes this visible without allowing Gemini's larger absolute memory scale to compress the other model cohorts. Exact measured-budget fit remains model-dependent.

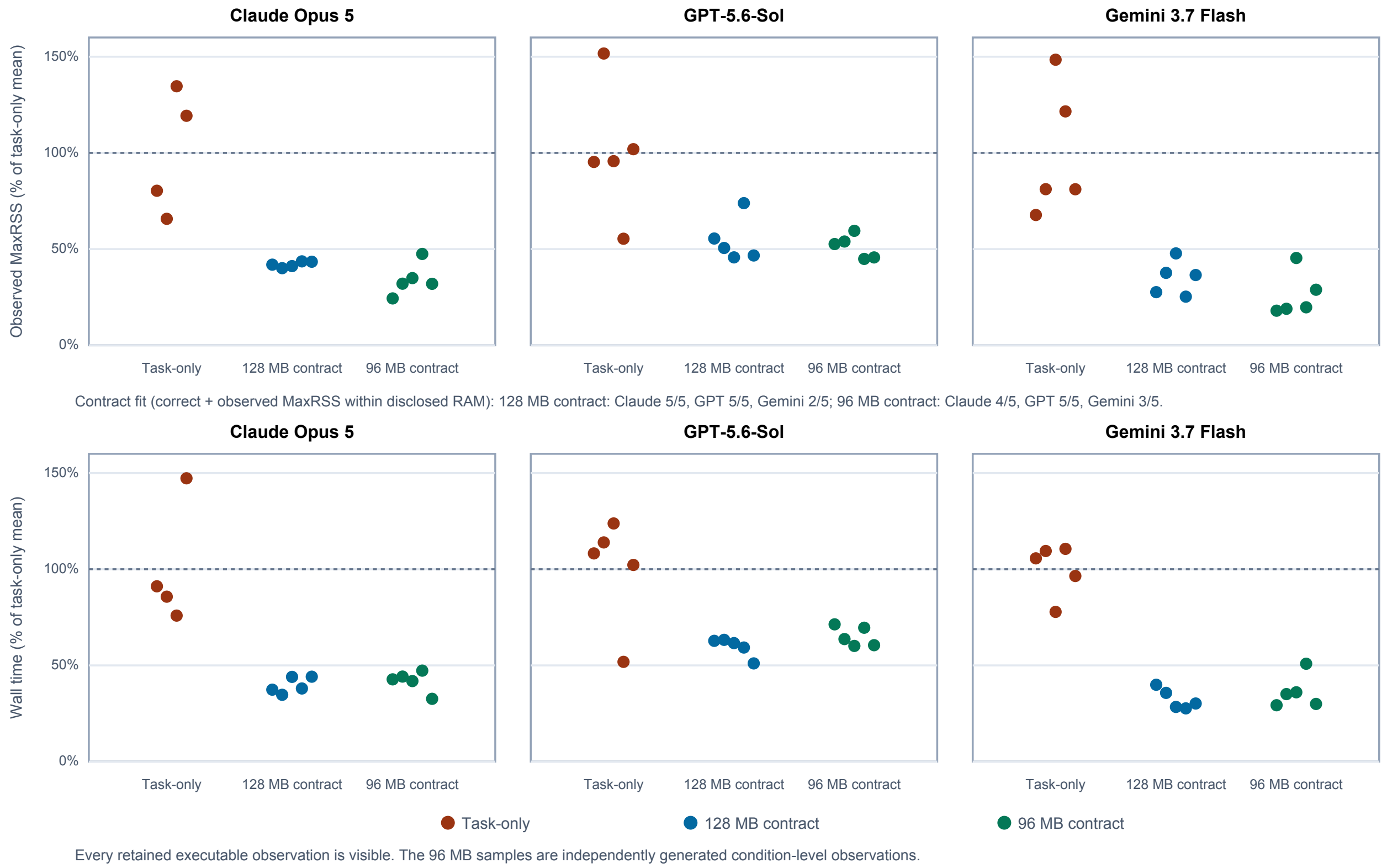

**Figure 2.** Every retained executable observation is indexed to its configured model's executable task-only-reference mean (100%). The upper row reports measured peak process memory and the lower row reports wall time. Table 2 provides the absolute means, percentage changes, and correct-and-within-budget counts; Appendix Figure A1 provides every raw MiB observation and the 96/128 MiB reference lines. The 96 MB programs are independently sampled condition-level observations.

## 5. Related work

Language-model agents increasingly combine reasoning with actions in external environments. ReAct, for example, interleaves reasoning traces and task-specific actions, using environment interaction to update action plans [4]. Software-agent evaluation has likewise made execution environments central: SWE-bench evaluates whether models can resolve real repository issues that require coordination with a codebase and its tests [5].

Code-efficiency benchmarks such as EffiBench and Mercury evaluate whether generated programs are not only correct but efficient in execution time and memory use [6, 7]. Execution-feedback approaches such as Reflexion and Self-Refine show how later feedback can improve subsequent generations [8, 9].

This work studies a complementary moment in the agent lifecycle. Rather than providing execution feedback after an action fails, substrate-aware planning supplies the relevant operating contract *before* an implementation is selected. The question is not whether execution feedback helps an agent repair a program; it is whether static execution context changes the program the agent chooses on its first generation. The controlled two-condition design makes that earlier implementation-selection effect directly inspectable.

## 6. The broader substrate-awareness agenda

The central finding is consequential: execution context that materially determines plan suitability belongs in an agent's planning state. The controlled result does not depend on a prescribed algorithm: a compact RAM/time contract alone shifted the generated implementation distribution. This is valuable even when a blind program happens to work under one environment, because suitability is defined by the environment in which deployment will actually occur.

The Python 3.9 runtime failure reported in Section 3.2 makes the same principle visible in software form. It is an illustrative observed compatibility incident, not a runtime-version-disclosure treatment. The controlled RAM/time experiment establishes the evidence in this paper; a future version-contract study can test whether supplying the runtime version changes first-pass compatibility.

The broader opportunity is substantial. A substrate-aware coding or tool-using agent can condition its plan on GPU memory, available CPU, runtime and dependency versions, tool latency, failure rates, permissions, quota, and cost. For agent harnesses, this makes pre-execution context a planning input rather than something the agent discovers only after a mismatch at runtime. The present result gives this agenda an empirical foundation: a minimal contract produces consequential changes in what the evaluated frontier configurations generate.

Across the 96 MB condition, the product of measured MaxRSS and wall time was 67-90% lower than the task-only cohort means. This is a duration-weighted observed process footprint, not a cloud-billing measurement. In containerized and serverless deployments, such a reduction can support tighter provisioning when it safely clears the relevant configured resource tier.

## 7. Research agenda and artifact availability

This paper establishes a controlled demonstration in a numerical-code setting. The next studies extend the same intervention to runtime-version-aware generation, accelerator-aware multimodal computation, constrained data pipelines, and dynamic tool telemetry. Each will test the same core principle against the operating dimensions that matter for its setting.

The evaluation archive is available at https://github.com/manu2/Context-Aware-Agent-Experiment [10]. It preserves the fresh direct-API manifest, prompts and dataset hashes, raw responses, generated scripts, numerical profiles, source-linked audit, and figure-generation code. The historical artifacts are retained for provenance and are not combined with the fresh cohort.

## Appendix A. Absolute resource profiles

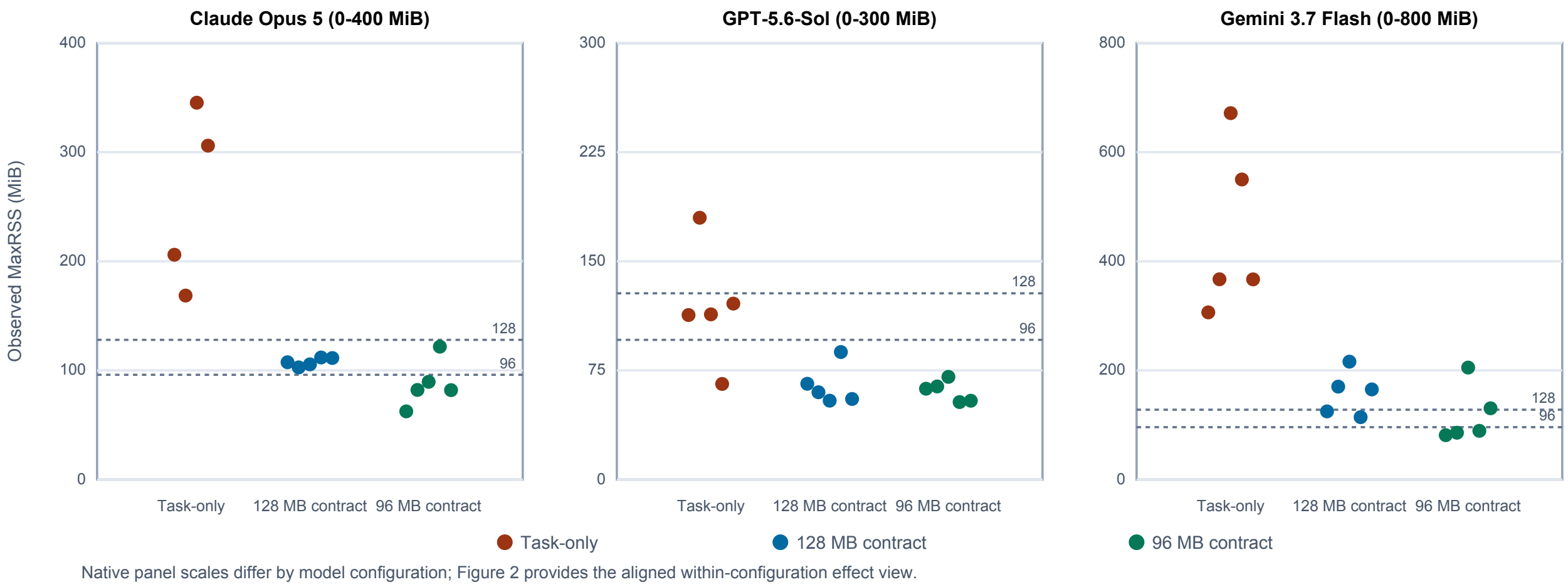

**Figure A1.** Every retained executable MaxRSS observation is shown in native MiB. Each panel uses its own MiB scale to reveal the within-configuration distributions; the panel heading states that scale. Dashed lines identify the prompt-labelled 96 MB and 128 MB boundaries, drawn at their corresponding 96 MiB and 128 MiB observed-RSS thresholds. Task-only and 128 MB contract-disclosed results provide reference distributions; the 96 MB programs are independently sampled condition-level observations.

## Appendix B. Reproducibility record

The frozen task-only prompt was:

```
Write a self-contained, executable Python script to process 'vectors.npy'
(containing an 8,000 x 1,024 float32 matrix). Compute the total sum of all
pairwise Euclidean distances between rows: sum_{i,j} ||v_i - v_j||_2 and print:
'TOTAL_DIST:<value>'. Constraint: Use ONLY numpy and standard library modules.
Do NOT import scipy or external packages.
```

The contract-disclosed prompt was the identical text followed by:

```
Execution environment:
RAM limit: 128 MB.
Execution time limit: 10.0 seconds.
```

No user-supplied system prompt or implementation hint was added. The archived manifest records the exact configured API IDs, output limits, and per-call timestamps. `gpt-5.6-sol` used temperature 1.0 and top-p 1.0 with an 8,192-token completion limit; `gemini-3.7-flash` used temperature 0.1 and top-p 0.95 with an 8,192-token output limit. `claude-opus-5` used an 8,192-token output limit and provider-default sampling because the configured API rejected explicit temperature and top-p controls. The artifact archive records the prompt and response for every included call, the deterministic dataset SHA-256, source hash, environment fingerprint, execution profile, and source-linked audit [10].

For trial-level inspection, the archive's machine-readable audit records every retained trial ID, condition, numerical result, exit status, MaxRSS, wall time, source hash, and observable structural features. The tables in this paper report the cohort summaries; the archived records provide the complete per-run evidence without compressing heterogeneous implementations into an artificial single strategy label [10].